\documentclass{article}

\usepackage[final]{bdu_2024}
\usepackage{natbib}

\usepackage[utf8]{inputenc} 
\usepackage[T1]{fontenc}    
\usepackage{hyperref}       
\usepackage{url}            
\usepackage{booktabs}       
\usepackage{amsfonts}       
\usepackage{nicefrac}       
\usepackage{microtype}      
\usepackage{xcolor}         

\usepackage{amsmath}
\usepackage{amssymb}
\usepackage{mathtools}
\usepackage{amsthm}
\usepackage{booktabs}
\usepackage{enumitem}

\title{Conformalised Conditional Normalising Flows for Joint Prediction Regions in time series}

\author{%
  Eshant English\thanks{primary contact: eshant.english@hpi.de \\The work was done during a research stay at UCI} \\
  Hasso Plattner Institute for Digital Engineering, Germany\\
  University of California, Irvine, USA \\
  \texttt{eshant.english@hpi.de} \\
  \And
  Christoph Lippert \\ 
  Hasso Plattner Institute for Digital Engineering, Germany \\
  Hasso Plattner Institute for Digital Health at the Icahn School of Medicine at Mount Sinai, USA\\
  \texttt{christoph.lippert@hpi.de} \\
}

\begin{document}

\maketitle

\begin{abstract}
Conformal Prediction offers a powerful framework for quantifying uncertainty in machine learning models, enabling the construction of prediction sets with finite-sample validity guarantees. While easily adaptable to non-probabilistic models, applying conformal prediction to probabilistic generative models, such as Normalising Flows is not straightforward. This work proposes a novel method to conformalise conditional normalising flows, specifically addressing the problem of obtaining prediction regions for multi-step time series forecasting. Our approach leverages the flexibility of normalising flows to generate potentially disjoint prediction regions, leading to improved predictive efficiency in the presence of potential multimodal predictive distributions.

\end{abstract}

\section{Introduction}
\label{Introduction}

Recent advances in Conformal Prediction have seen its application extended to diverse domains, including Gaussian Processes \citep{jaber2024conformal} and Monte-Carlo methods \citep{bethell2024robust}, driven by the attractive finite-sample validity guarantees offered via conformal prediction \citep{vovk2005algorithmic,angelopoulos2022gentle}. However, direct application to generative models, such as Normalising Flows \citep{Kobyzev_2021,papamakarios2021normalizing} or Diffusion models \citep{song2021scorebasedgenerativemodelingstochastic, luo2022understanding}, has been limited due to their primary focus on generation rather than prediction tasks.  

Notably, discrete-time normalising flows offer a unique advantage over other generative models: simulation-free exact likelihood computation. This enables direct density evaluation at a given sample, naturally extending to conditional density computation. In this work, we exploit this property of normalising flows to derive prediction regions for multi-step time series forecasting.

A prediction region is defined as a volume in $H$-dimensional space, where $H$ is the forecast horizon, capturing the true forecast with probability $1-\epsilon$.  Constructing such regions for time series is challenging due to temporal dependencies. Existing methods, such as \cite{stankeviciute_conformal_2021}, rely on Bonferroni corrections based on conditional independence assumptions on the errors, leading to overly conservative predictions as $H$ increases. Moreover, these approaches typically yield continuous band or rectangular regions, which may not be optimal for multimodal predictive distributions present in irregular time series.

Traditional conformal methods often rely on univariate conformity scores, with residuals being a natural choice in regression settings. However, this approach does not readily generalise to multivariate prediction cases. We propose the use of conditional density as a conformity score, capitalising on its adaptability to the historical context and its ability to quantify the plausibility of a given prediction naturally.

In summary, our key contribution lies in extending conformal methods to conditional normalising flows, thereby enabling adaptive and efficient predictive inference for multi-step time series forecasting (or more generally multivariate output).

\section{Related works}
Previous research has explored normalising flows for time series forecasting \citep{feng2023multiscale,rasul2021multivariate}. However, existing methods for uncertainty quantification in this context either lack coverage guarantees or rely on unrealistic assumptions when dealing with multi-step or multivariate forecasts. In the univariate case, conformalisation using residuals is straightforward. However, for higher-dimensional predictions, mapping residuals to quantiles is not trivial.

Existing approaches attempt to address this issue through various means, such as transforming multivariate residuals into Euclidean distances \citep{messoudi2021copulabased}, using Chebyshev distance after modulation \citep{english2024janetjointadaptivepredictionregion, zhou2024conformalizedadaptiveforecastingheterogeneous}, or modelling interdependencies between successive time steps using copulas \citep{sun2024copula}. Nevertheless, these methods often result in large, continuous prediction regions that may be inefficient for capturing multimodal predictive distributions in time series.  Achieving predictive efficiency, where the volume of the prediction set is minimised, becomes a crucial objective in such scenarios.

A recent work \cite{colombo2024normalizingflowsconformalregression} also combined Normalising Flows with Conformal Prediction. However, they do not use Normalising Flows to obtain density estimates for conformal scores; instead, they train the calibration process with Normalising Flows acting on the joint distribution of the errors. To the best of our knowledge, no existing conformal method for time series forecasting can produce multiple disjoint regions when the predictive distribution is multimodal. Building upon prior work, we propose a novel approach applicable to scenarios with multiple independent time series. Our method aims to overcome the limitations of previous approaches and provide a more efficient and flexible way to quantify uncertainty in time series forecasting.
\section{Preleminaries}

\subsection{Inductive Conformal Predictors}
Inductive conformal predictors provide a computationally efficient alternative to Full conformal predictors for constructing prediction intervals by training the model only once. Given a dataset $\mathcal{D} = \{(x_i, y_i)\}_{i=1}^n$ drawn from a distribution $P$, we divide it into a training set $\mathcal{D}_{\text{train}}$ of size $m$ and a calibration set $\mathcal{D}_{\text{cal}}$ of size $l$, where $n = m + l$.

A point predictor $h$ is trained on $\mathcal{D}_{\text{train}}$, and a conformity measure $A$ is used to assess the \textit{conformity} of an example relative to the calibration set. We compute conformity scores (denoted by $\alpha$) for both the calibration set and a test point $x^*$ with a postulated label $y \in \mathcal{Y}$, the label space:
\begin{itemize}[label=$\circ$]
    \item $\alpha_i = A(h, \mathcal{D}_{\text{train}}, (x_{m+i}, y_{m+i}))$, for  $i = 1, \dots, l$
    \item $\alpha^* = A(h, \mathcal{D}_{\text{train}}, (x^*, y))$
\end{itemize}

The prediction set ($\Gamma^\epsilon$) at significance level $\epsilon$ is then defined as:
\begin{align}
\label{equation1}
\Gamma^\epsilon(x^*) = \left\{y: \frac{|\{i: \alpha_i \leq \alpha^*\}| + 1}{l + 1} > \epsilon \right\}.
\end{align}

This set represents the values of $y$ that are deemed \textit{conforming} compared to the calibration set, with a probability of at least $1-\epsilon$. Instead of comparing all the conformity scores, one can invert the quantile function of the conformity scores on the calibration set to get a threshold conformity score.

\subsection{Conditional Normalising Flows}

Conditional normalising flows (CNFs) are normalising flows that incorporate additional conditioning information, $c$, into the transformation. This allows the learned distribution to be adapted based on the context provided by $c$. Formally, a CNF is a function $f: \mathbb{R}^D \times \mathbb{R}^M \rightarrow \mathbb{R}^D$, where $M$ is the dimensionality of the conditioning variable. 

Given a dataset $\mathcal{D} = \{(x_1, c_1), \dots, (x_n, c_n)\}$, we aim to maximise the conditional log-likelihood:
$$\max_{f} \sum_{i=1}^n \log p_{X|C}(x_i | c_i),$$
where $p_{X|C}$ is the conditional probability density function of the data given the conditioning variable, $c$. Using the change of variables formula, we express the conditional density as:
\begin{align}
  p_{X|C}(x|c) = p_{Z|C}(f^{-1}(x, c) | c) \left|\det J_{f^{-1}}(x, c)\right|.
\end{align}
Here, $J$ refers to the Jacobian of the transformation and $\det$ refers to the determinant. Then, the maximum log-likelihood optimisation objective can be stated as:
$$ \max_{f} \sum_{i=1}^n \left[ \log p_{Z|C}(f^{-1}(x_i, c_i) | c_i) + \log \left|\det J_{f^{-1}}(x_i, c_i)\right| \right].$$
Similar to standard normalising flows, the first term can be computed analytically by assuming a simple form for $p_{Z|C}$ (e.g., a Gaussian distribution with mean and variance dependent on $c$). The second term, involving the Jacobian determinant, requires careful architectural choices for efficient computation.

CNFs are typically constructed as a composition of simpler invertible transformations, each of which can be conditioned on $c$. Various techniques, such as affine coupling layers or autoregressive flows, are easily adaptable to the conditional setting. The choice of transformations and their conditioning mechanism influences the expressivity and flexibility of the CNF model.

\section{Conformalised Conditional Normalising Flow for Joint Prediction Region in time series (CCN-JPR)}

We consider the problem of probabilistic forecasting for a collection of $n$ independent multivariate time series, each denoted by $\mathbf{x}^{(i)} \in \mathbb{R}^{D \times (T+H)}$, where $T$ is the length of the context provided to the model and $H$ is the forecast horizon. We partition the time series' into a training set proper of $m$ series and a calibration set of $l$ series, such that $n = m + l$. 

Let $\mathbf{x}_t^{(i)} \in \mathbb{R}^{D \times (T+H)}$ represent the $i$-th time series at time step $t$. Given the context window $\mathbf{x}_{1:T}^{(i)}$, the conditional distribution of future values $\mathbf{x}_{T+1:T+H}^{(i)}$ over the entire dataset $ \mathbf{x} = \{\mathbf{x}_{1:T+H}^{(i)}\}_{i = 1}^{m}$ is modeled as:
$$
p(\mathbf{x}_{T+1:T+H}^{} | \mathbf{x}_{1:T}^{}; \theta) = \prod_{i=1}^{m} p(\mathbf{x}_{T+1:T+H}^{(i)} | \mathbf{h}_T^{(i)}; \theta),
$$
where $\mathbf{h}_T^{(i)}$ is temporal summarisation for the $i$-th time series at time step $T$, and $\theta$ represents the model parameters.

Similar to \cite{rasul2021multivariate}, it is possible to use an RNN-based architecture or a transformer to summarise the temporal history instead of flow transformation to incorporate it. For RNNs, $\mathbf{h}_T^{(i)}$ could be the hidden state of the RNN for the $i$-th time series at time step $T$. For any arbitrary time step $t$, the hidden state is updated recursively:
$$
\mathbf{h}_t^{(i)} = \text{RNN}(\mathbf{x}_{t-1}^{(i)}, \mathbf{h}_{t-1}^{(i)}),
$$
To capture the potentially complex, high-dimensional distribution of $\mathbf{x}_{T+1:T+H}^{(i)}$, we use a flow to model the emission distribution $p(\mathbf{x}_{T+1:T+H}^{(i)} | \mathbf{h}_{T}^{(i)}; \theta)$:
$$
\mathbf{x}_{T+1:T+H}^{(i)} = g_{\theta}(\mathbf{z}_{T+1:T+H}^{(i)}; \mathbf{h}_T^{(i)}),
$$
where $g_{\theta}$ is an invertible transformation parameterized by $\theta$, and $\mathbf{z}_{T+1:T+H}^{(i)}$ is a latent variable drawn from a simple base distribution (e.g., standard Gaussian). 

Our specific implementation utilises a stack of $K$ conditional affine coupling layers for $g_{\theta}$, each conditioned on the hidden state $\mathbf{h}_T^{(i)}$ (identity transformation on $\mathbf{x}_{1:T}^{(i)}$ for the experiments) to accommodate the temporal dynamics. The model is trained via maximum likelihood estimation using stochastic gradient descent, and forecasts are generated by sampling from the conditional distribution.

Once a model is trained using the training set proper, we can use the calibration set to calculate the conditional density around the true prediction given the historical context by using the change of the variable formula as stated in Equation 2. We consider this as our conformity score and can apply Equation~\ref{equation1} for any postulated label to know if it is a part of the prediction set. Additionally, the regions formed are adaptive to the historical context by using conditional density as a conformity score.

We can also sort all the conformity scores and obtain the desired quantile $q_{\epsilon}$ of the sorted scores. To construct a region that contains the true prediction with the desired significance level $\epsilon$, we can, then, sample points from the multi-variate label space either through a grid or through other methods such as Monte-Carlo sampling, the sampled points with a conditional density larger than $q_{\epsilon}$ form cluster(s) which can be several if there are multiple modes of the predictive distribution. Unlike other baselines, a region formed like this has no geometrical restrictions.

\section{Empirical Results}
We assess the \textit{coverage} of our model on two synthetic time-series datasets and one real-world dataset (COVID-UK) from \cite{kipf2018neural}, following the experimental protocol of \cite{sun2024copula}. 

For both \textit{synthetic} datasets, we forecast 24 steps ahead ($H=24$) using a history of 35 steps ($T=35$), Each time step consists of two-dimensional data ($\mathbb{R}^2$).  The datasets differ in the magnitude of the added Gaussian noise: $\sigma=0.01$ and $\sigma=0.05$, respectively. For the COVID-UK dataset, which is a difficult time series due to irregularity, we predict daily cases for the upcoming 10 days based on the preceding 100 days.

\begin{table}[!ht]
\centering
\caption{Comparison of the coverages (nominal) on the test set for $\epsilon=0.1$. The coverage values closer to 0.9 are highlighted in bold on each dataset. Our method achieves values close to 0.9 whereas CopulaCPTS always undercovers, MC-dropout severely undercover, and CRNN often overcovers (except on the COVID-UK dataset).}
\label{tab:combined_coverage}

\begin{tabular}{@{}l lc c c@{}}  
\toprule
&\textbf{Method} & \textbf{Particle1} & \textbf{Particle5} & \textbf{COVID-UK} \\ 
\midrule

& MC-Dropout          & 0.79 & 0.43 & 0.00 \\
&CRNN       & 0.99 & 0.93 & \textbf{0.87}\\
&CopulaCPTS  & 0.86 & 0.86 & 0.78\\
&\textbf{CCNF-JPR} & \textbf{0.91} & \textbf{0.89} &\textbf{0.87}  \\ 
\bottomrule
\end{tabular}
\end{table}

We computed the coverages by checking if the true value is accepted by the conditional density threshold used as a conformity score by our method. Table~\ref{tab:combined_coverage} presents results for all the datasets: ($\sigma=0.01$) \textit{particle1} and ($\sigma=0.05$) \textit{particle5} dataset. Our proposed method demonstrates coverage closest to the nominal 0.9 level ($\epsilon$ = 0.1). CRNN overcovers(except on the COVID-UK dataset), and other methods, notably MC-dropout, exhibit severe undercoverage.

\section{Discussion and Future Work}
In this work, we introduced a novel approach to conformalise conditional normalising flows for multi-step time series forecasting, enabling the construction of prediction regions without assuming specific geometric shapes like spheres, bands, or ellipses \citep{cleaveland2024conformal,xu2024conformal}. By conditioning the conformity scores on the time series history, our method also yields adaptive prediction regions, a feature not commonly found in existing approaches (\cite{english2023kernelised, zhou2024conformalizedadaptiveforecastingheterogeneous} being notable exceptions, but can only have rectangular regions). Although our work focussed on time series forecasting, the idea is trivially applicable to multivariate regression as multivariate regression can be considered a special case where the time steps/indices are independent. One bottleneck is that exhaustive sampling might be computationally expensive if the label space is high-dimensional. Efficient sampling schemes can be explored for computational feasibility.

While normalising flows are often employed with neural networks, this might not always be desirable when using classical statistical methods such as ARMA models. Future research could explore extending existing kernel-based flow approaches \citep{meng2020gaussianization,english2023kernelised} to conditional normalising flows (conditioned on prediction from the classical methods), potentially mitigating these challenges.

A further limitation is the lack of standard datasets exhibiting multimodal predictive distributions in time series, which are necessary to showcase the full potential of our approach. Future work will focus on addressing these limitations and conducting comprehensive empirical evaluations.

\subsubsection*{Acknowledgements}
We extend our gratitude to Thomas Gaertner for valuable feedback on an earlier draft and to Prof Vladimir Vovk for helpful discussions. This research was funded by the HPI Research School on Data Science and Engineering.

\bibliography{neurips_2024}
\bibliographystyle{plainnat}
\end{document}